\PassOptionsToPackage{round,authoryear}{natbib}
\PassOptionsToPackage{table,dvipsnames}{xcolor}
\documentclass[]{style/company_light}

\usepackage[utf8]{inputenc}
\usepackage{CJKutf8}
\usepackage{hyperref}
\usepackage{url}
\usepackage{booktabs}
\usepackage{amsfonts}
\usepackage{amsmath}
\usepackage{amsthm}
\usepackage{microtype}
\usepackage{graphicx}
\usepackage{placeins}
\usepackage{flafter}
\usepackage{multirow}
\usepackage{enumitem}
\usepackage{adjustbox}
\usepackage{array}
\usepackage{tabularx}
\usepackage{longtable}
\usepackage{amssymb}
\usepackage{multicol}
\usepackage{xcolor}
\usepackage{xspace}
\usepackage{tikz}
\usetikzlibrary{arrows.meta,positioning,fit,calc,backgrounds}
\usepackage{algorithm}
\usepackage{algpseudocode}
\floatname{algorithm}{Algorithm}
\newtheorem{definition}{Definition}
\ifdefined\pdfsuppresswarningpagegroup\pdfsuppresswarningpagegroup=1\fi

\newtoggle{hldraft}
\toggletrue{hldraft}

\newcommand{\hlnumtbdlist}{}

\newcommand{\defval}[2]{\csdef{hlnum@#1}{#2}\csundef{hlntbd@#1}}
\newcommand{\defTBD}[2]{%
  \csdef{hlnum@#1}{#2}\csdef{hlntbd@#1}{}\listadd{\hlnumtbdlist}{#1}}

\newcommand{\val}[1]{%
  \ifcsdef{hlnum@#1}{%
    \ifcsdef{hlntbd@#1}{%
      \iftoggle{hldraft}{\textcolor{brandaccent}{\bfseries\csuse{hlnum@#1}}}%
        {\PackageError{hlnumbers}{Unmeasured key '#1' used with hldraft off}%
          {Key '#1' is still \string\defTBD. Measure it and convert it to
           \string\defval, or re-enable \string\toggletrue{hldraft}.}%
         \csuse{hlnum@#1}}%
    }{\csuse{hlnum@#1}}%
  }{%
    \textcolor{red}{\bfseries[?\,\texttt{#1}]}%
    \PackageWarning{hlnumbers}{Undefined number key '#1'}%
  }%
}

\newcommand{\hlnumtbditem}[1]{\item \texttt{#1} \dotfill\ \csuse{hlnum@#1}}
\newcommand{\printTBD}{%
  \iftoggle{hldraft}{\ifdefempty{\hlnumtbdlist}{}{%
    \section*{Outstanding measurements}
    Placeholders still standing in for unmeasured quantities. This list must be
    empty before submission.
    \begin{itemize}[nosep,leftmargin=*]
      \forlistloop{\hlnumtbditem}{\hlnumtbdlist}
    \end{itemize}}}{}}

\newcommand{\hlnumtbdlog}[1]{\typeout{  #1 = \csuse{hlnum@#1}}}
\AtEndDocument{%
  \iftoggle{hldraft}{%
    \typeout{^^J===== OUTSTANDING MEASUREMENTS (draft mode) =====}%
    \forlistloop{\hlnumtbdlog}{\hlnumtbdlist}%
    \typeout{===== end =====^^J}%
  }{}}

\defval{ntasks}{25}
\defval{ncategories}{six}
\defval{nbackends}{four}
\defval{nsettings}{two}
\defval{nenvs}{212}
\defval{ntemplates}{32}
\defval{ngroups}{seven}
\defval{mainhours}{72}
\defval{ntracks}{two}
\defval{ntrackenvs}{400}
\defval{npersontypes}{eight}
\defval{npumps}{five}
\defval{npens}{10}
\defval{ninsulins}{10}
\defval{nsensors}{eight}
\defval{nmeters}{six}
\defval{nprofiles}{seven}
\defval{nclinical}{seven}
\defval{ntests}{253}
\defval{horizonhours}{24}
\defval{gridminutes}{5}

\defval{refnone}{0.42}
\defval{refpatient}{0.82}
\defval{reforefzero}{0.56}
\defval{refbone}{0.82}
\defval{refbtwo}{0.84}
\defval{refbthree}{0.85}
\defval{refbfour}{0.86}
\defval{refcausal}{0.90}
\defval{refomni}{0.98}
\defval{refassistedalone}{0.79}
\defval{refassistedcausal}{0.90}
\defval{refassistedomni}{0.99}
\defval{refloopbest}{0.94}
\defval{refloopcausal}{0.93}
\defval{refnoloopbest}{0.77}
\defval{refnoloopcausal}{0.87}
\defval{headroomnoloop}{0.10}
\defval{assistedgain}{0.110}
\defval{assistedgainlo}{0.073}
\defval{assistedgainhi}{0.144}
\defval{hardcertified}{169}
\defval{hardcandidates}{six}
\defval{hardcausalmin}{0.95}
\defval{hardordinarymax}{0.6}
\defval{tiereasy}{66}
\defval{tiermedium}{67}
\defval{tierhard}{67}

\defval{nmodels}{five}
\defval{bestmodel}{Claude Opus~5.5}
\defval{bestscore}{73.54}

\newcommand{\fwname}{\mbox{DiabAgent}\xspace}
\newcommand{\apcname}{Agentic Physiological Control\xspace}
\newcommand{\apc}{\mbox{APC}\xspace}
\newcommand{\simname}{\mbox{DiabetSim}\xspace}
\newcommand{\benchname}{\mbox{SweetBench}\xspace}

\newcommand{\code}[1]{\texttt{#1}}

\title{DiabAgent: Toward Agentic Physiological Control}
\author[1,*]{Chengxi Li}
\author[1,*]{Shaoxuan Li}
\author[1,*]{Deyao Hong}
\author[1]{Qingle Liu}
\author[1]{Yi Huang}
\author[1]{Puheng Bian}
\author[1]{Minzhi Xu}
\author[1]{Lanyue Zhang}
\author[1]{Mingze Sun}
\author[2]{Xin Li}
\author[3]{Calvin Xiao}
\author[1]{Shengyuan Wang}
\author[1]{Huan-ang Gao}
\author[1]{Jiazheng Xu}
\author[1,\dagger,\ddagger]{Kaisen Yang}
\author[1,\dagger]{Hongning Wang}
\affiliation[1]{Tsinghua University}
\affiliation[2]{Sweetie Online Community}
\affiliation[3]{Einsia AI}
\contribution{$^{*}$ Core contributor; $^{\dagger}$ Corresponding author; $^{\ddagger}$ Project leader.}
\metadata[Contact]{\email{yks23@mails.tsinghua.edu.cn}}
\metadata[Homepage]{\href{http://diabagent.net9.org/}{diabagent.net9.org}}
\metadata[Codebase]{\href{https://github.com/THU-CST-SAST/DiabAgent}{github.com/THU-CST-SAST/DiabAgent}}

\abstract{As pretraining scales, medical language models have advanced rapidly in clinical question answering.
Whether they can achieve effective physiological control through sustained intervention has not been systematically evaluated.
Type~1 diabetes provides a concrete setting: impaired insulin secretion creates dependence on exogenous insulin, with dosing continually adjusted using information about meals, exercise, illness, and glucose measurements.
We introduce \fwname, a research suite comprising the \apcname (\apc) framework, the \simname simulator, and the \benchname benchmark.
\apc controls glucose through medical-device tools, using ongoing observations to infer individual states, accumulate personal experience, and choose treatment actions.
We build \simname to simulate different patients, events, and device conditions for scalable virtual testing and training.
Using \simname, we construct \benchname by instantiating \val{ntemplates} scenario designs as \val{nenvs} glucose-control tasks.
With Claude Opus~5.5 and a fixed decision interval of one hour, \apc achieves a COGI of 73.5, compared with 71.2 for the open-source closed-loop baseline OpenAPS.
Ablations further show that individual adaptation and contextual information improve glucose control.
Combining high-frequency rule-based control with lower-frequency agent decisions shows potential for balancing control performance and computational cost.
Our work provides a framework, simulator, and evaluation foundation for agents to participate in human physiological control.}

\begin{document}

\maketitle

\section{Introduction}
\label{sec:introduction}

As pretraining scales, medical language models have advanced rapidly in clinical question
answering and treatment advice~\citep{multimedqa,healthbench}. Agents combine language
models with tools, memory, and environment feedback to execute multi-step
tasks~\citep{react2023,reflexion2023}. They complete long tasks in software and service
environments and reuse experience in later decisions~\citep{sweagent2024,taubench,expel2024}.
Medical agents conduct diagnostic dialogue, analyze
treatment options, and operate electronic health
records~\citep{amie2025diagnosis,gao2025txagent,medagentbench}. Evaluation of these tasks
primarily concerns responses or operations on records. Assessing whether language models
can control physiology through sustained intervention requires systematic evaluation of
the resulting physiological trajectories.

Type~1 diabetes provides a concrete setting for this problem. People with type~1 diabetes
have impaired insulin secretion and depend on exogenous insulin for glucose control.
Dosing needs vary with meals, exercise, illness, and individual states, while insulin
continues to act for hours, linking current decisions to earlier treatment and subsequent
events~\citep{holt2021consensus,riddell2017exercise}. Differences between people, daily
changes in insulin needs, and delayed, noisy, or missing glucose measurements make
control a long-horizon task requiring ongoing state inference, treatment adjustment,
and feedback tracking. Automated insulin delivery and learned closed-loop controllers
have advanced glucose control~\citep{brown2019closedloop,sherr2022aid,hettiarachchi2024g2p2c,sarkar2026llmt1d}.
Existing dynamical models and evaluation metrics also provide a basis for simulating
glucose dynamics and assessing control outcomes~\citep{deichmann2023,tirconsensus}.
Together, these conditions support systematic study of how agents combine contextual
information, personal experience, and device tools for sustained physiological control.

Rule-based closed-loop control typically organizes dosing around structured glucose
measurements, insulin records, carbohydrate estimates, and prescribed therapy
parameters~\citep{openaps}. Using descriptions of a changed meal plan, illness recovery,
or a device problem requires mapping them to predefined inputs and response rules.
This mapping constrains how the controller can use cues in free-text context and personal
treatment notes. Interacting rare events and individual differences also require judging
which cues explain current glucose changes and subsequent dosing
needs~\citep{holt2021consensus}. \textbf{Can an agent interpret this context and personal experience, act through medical-device
tools, and translate that understanding into better physiological control?} This question motivates the \apcname (\apc) framework
that we study in \fwname.

We evaluate \apc with five models. With Claude Opus~5.5, it achieves a COGI of 73.5,
compared with 71.2 for OpenAPS. Further experiments show that records and review support cross-day improvement
without weight updates, and that added context can improve control more than increasing
agent decision frequency in the tested cases. More frequent agent calls also increase
computation. Case studies that switch a background closed-loop (CL) controller on and off
indicate the potential of combining frequent background glucose regulation with less
frequent agent decisions that integrate observations and personal experience.

\begin{figure}[!htbp]
\centering
\includegraphics[width=\linewidth]{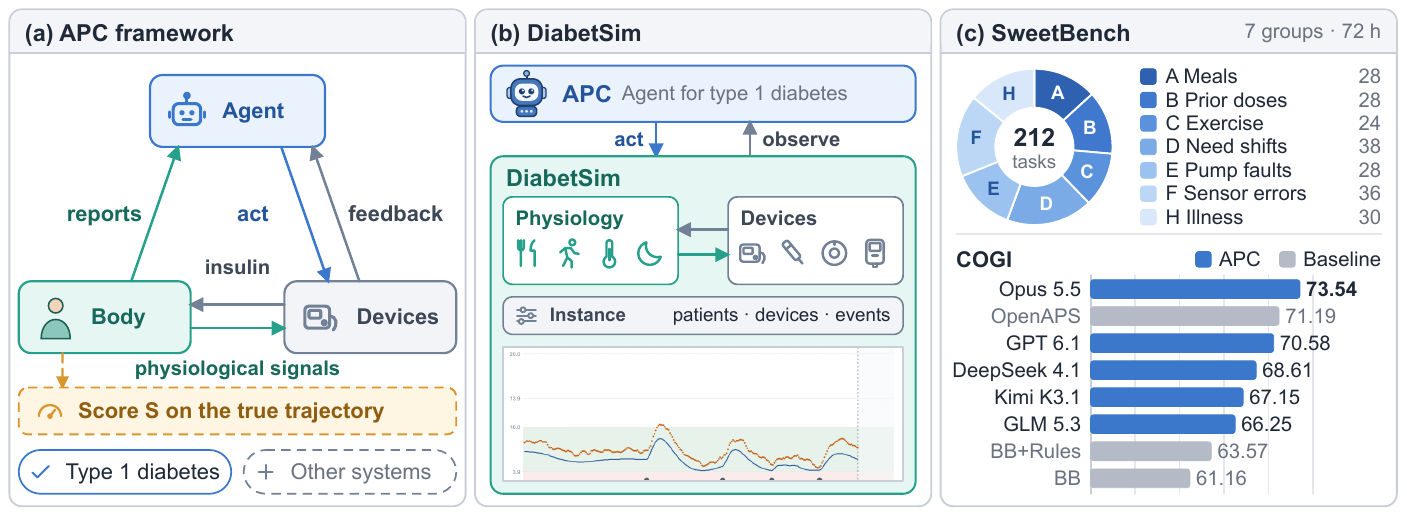}
\caption{\textbf{\apc and its first instantiation.} (a) The \apc loop, in which an agent
acts on a physiological system through device tools and is scored on the physiological
outcome. This work applies it to type~1 diabetes, and other systems remain open. (b) \simname
combines a physiological engine, device models, and an instance generator; \apc acts through
the devices and observes their readings. The plot shows a simulated glucose trajectory.
(c) \benchname composes \val{nenvs} tasks in \val{ngroups} scenario groups and scores the
true glucose trajectories, which are hidden from the agent. The donut chart gives the tasks
per group, and the bars show the mean COGI of \apc with five backend models and of the
three baselines.}
\label{fig:hero}
\end{figure}

We make three contributions:

\begin{enumerate}[leftmargin=*]
  \item We introduce \textbf{\apc}, an agentic control framework that treats human
    physiology as an interactive environment, and apply it to glucose control. It acquires information
    through monitoring devices, delivers medication through medical devices, and adjusts
    decisions from feedback to optimize glucose-control metrics.
  \item We build \textbf{\simname}, a configurable glucose-control simulator that
    integrates individual physiology, medical devices, events, and observation conditions.
    It reconstructs the interaction environment faced by \apc and provides a foundation
    for reproducible virtual testing.
  \item We construct \textbf{\benchname}, a benchmark of \val{nenvs} glucose-control tasks.
    \apc with Claude Opus~5.5 achieves a COGI of 73.5, compared with 71.2 for OpenAPS. Further ablations show improved
    control over days with personal records and end-of-day review.
\end{enumerate}

\section{\apcname}
\label{sec:problem}

\subsection{The \apc Framework}
\label{sec:apc}
\label{sec:setting}
\label{sec:requirements}

\apc treats treatment as a control problem in which the controller is an agent and the
sensors and actuators are medical devices.

\begin{definition}[\apcname]
An \apc problem consists of a physiological system with hidden state, medical devices that
measure and act on it, reports from the person, and a score on the physiological trajectory.
An \apc agent repeatedly observes, acts through the devices, and decides when to reassess,
so as to maximize the expected score.
\end{definition}

The state $x_t$ comprises physiology, device status, and context, and evolves under
individual parameters $\theta$. Let $d$ denote the device configuration and $\rho$ the
reporting configuration, which specifies what the person reports, when, and how
accurately. Then
\[
  x_{t+1}\sim P_{\theta}(\cdot\mid x_t,a_t),
  \qquad o_t\sim O_{d,\rho}(\cdot\mid x_t),
  \qquad a_t\in\mathcal{A}_{d}(h_t).
\]
This is a partially observable Markov decision process \citep{kaelbling1998pomdp} in which
the devices determine both the observation model and the action set. The agent keeps the
visible history $h_t=(o_{\le t},a_{<t})$ and a personal record $M_t$ summarizing past
interaction with the same individual, and selects actions with a policy
$\pi(a_t\mid h_t,M_t)$. The score is computed from the true physiological variables
$m_t=f(x_t)$, which the agent never observes directly. Over a horizon $T$, the objective is
$\max_{\pi}\mathbb{E}_{\pi}[S(m_{0:T})]$.

\begin{figure}[!htbp]
\centering
\includegraphics[width=\linewidth]{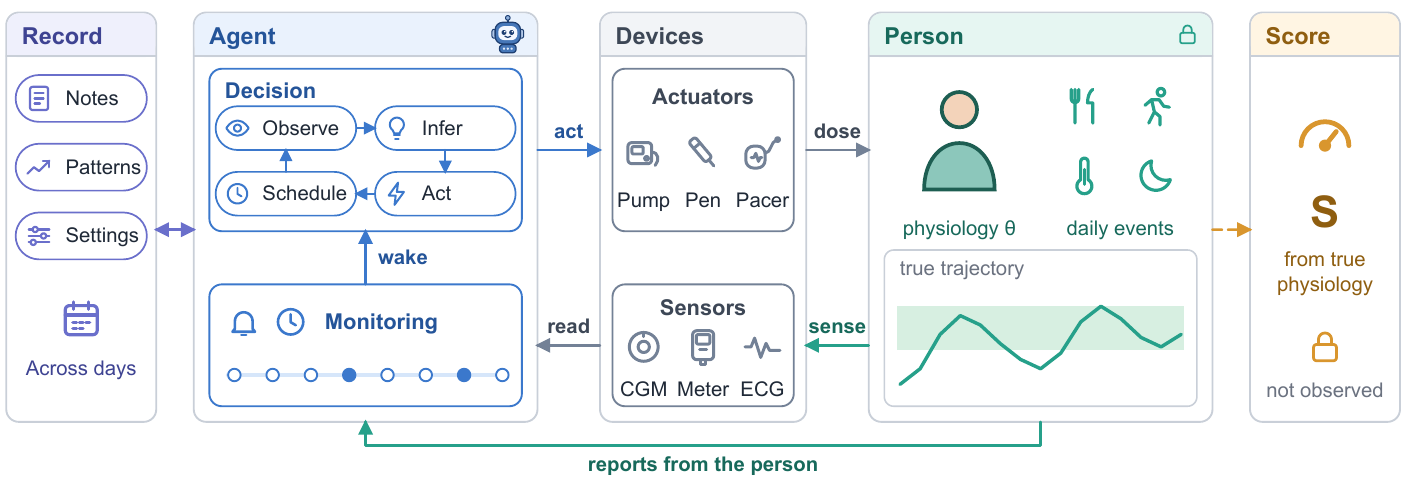}
\caption{\textbf{The \apc loop.} The agent reaches the physiological system only through
medical devices. Sensor readings and the person's reports return through monitoring, which
wakes the agent at its chosen time or when a trigger fires, and the personal record persists
across decisions and days. The score $S$ is computed from the true physiological trajectory.}
\label{fig:loop}
\end{figure}

An \apc agent has four components that close this loop (Figure~\ref{fig:loop}).
Decision making selects an intervention, a measurement, or waiting, and states its
rationale. Device tools execute the operation and return readings and delivery
logs. Monitoring watches incoming observations and returns control to the agent
when its chosen waiting interval ends or a specified condition occurs.
The personal record carries what has been learned about the individual into later
decisions.

The design addresses three requirements. Interpretability makes the basis of
an action traceable through decision rules or an explicit rationale.
Information integration combines glucose measurements, natural-language and multimodal reports,
and personal history to infer the hidden state and select interventions.
Long-tail event adaptation uses contextual reports to adjust treatment
when events overlap or fall outside predefined response categories.

To instantiate \apc in a new domain, one specifies the physiological variables and the score
$S$, the devices with their observation models and operations ($O_{d,\rho}$,
$\mathcal{A}_d$), the reports the person can provide, and an environment in which agents can
be evaluated safely. Section~\ref{sec:diabetes-control} makes these choices for type~1
diabetes, and Sections~\ref{sec:environment} and~\ref{sec:benchmark} build the environment
and the tasks.

\subsection{Glucose Control}
\label{sec:diabetes-control}
\label{sec:self-evolving}

Rule-based glucose controllers organize dosing around structured glucose and therapy
records~\citep{holt2021consensus,openaps}. LLM-T1D distills a reinforcement-learning
policy into a language model that doses insulin and explains its
decisions~\citep{sarkar2026llmt1d}.

We instantiate \apc for glucose control in type~1 diabetes, combining device observations
with contextual reports and personal experience. The state comprises glucose, insulin
still acting, carbohydrate absorption, and device
status; meals, activity, and illness perturb this state. \apc observes sensor
or meter readings, delivery logs and alarms, and reports of meals, activity, and device
problems. Its actions are insulin delivery through boluses, temporary basal rates,
or injections, measurement requests such as scans and fingersticks, device management such as
cancellation, suspension, and resumption, and waiting. The score combines time in
range, hypoglycemia, and variability of true glucose (Section~\ref{sec:scoring};
\citealp{tirconsensus}). The idealized reference is a healthy pancreas, which regulates glucose continuously without
devices or reports.

At each decision, \apc reviews recent readings, insulin still acting, reported events,
and its personal record, and estimates where glucose is heading. It then chooses one
operation supported by the current devices and states why. The same decision fixes when to
look again and which events, such as a new reading, a reported meal, or an alarm, should wake
it earlier. Monitoring holds control until one of these occurs.

The personal record stores notes on how this individual has responded to meals, activity,
and insulin. In multi-day episodes the conversation resets each day while the record
persists, and revised therapy settings persist where the experimental condition allows.
Adaptation therefore happens through experience with model weights fixed, and
Section~\ref{sec:exp-evolve} evaluates the effect of keeping the record.
Appendix~\ref{app:interface} specifies the observation and action interface.

\section{\simname}
\label{sec:environment}
\label{sec:simulator}

\simname is a configurable environment for virtual testing of glucose-control agents.
It comprises a physiological engine, monitoring and delivery device models, and an
instance generator. Together, these components simulate individual glucose responses,
everyday events, and device-mediated interventions. Agents observe device readings and
event reports; the evaluator records the underlying glucose trajectory for scoring.

\subsection{Simulator Engine}
\label{sec:person}

The engine models glucose--insulin dynamics using \citet{deichmann2023}, including the
effects of physical activity. Individual parameters $\theta$ specify insulin sensitivity
and body-mass-dependent responses. The simulator includes \val{npersontypes} physiological
types with different body masses and dawn phenomena, together with day-to-day variation
in insulin sensitivity. Meals introduce carbohydrate absorption, while activity and
illness modify subsequent physiological responses. The simulator configures therapy settings
separately from physiological parameters, allowing prescribed treatment to differ from
an individual's insulin requirements. An event-driven scheduler advances physiological
and device states as external events occur and treatment actions execute.

\subsection{Devices}
\label{sec:world}
\label{sec:devices}

\textbf{Observation.} Monitoring devices determine the measurements available to the agent.
The simulator includes \val{nsensors} sensor models and \val{nmeters} glucose-meter models,
parameterized by published accuracy specifications \citep{vettoretti2019g6,iso15197}.
Real-time continuous glucose monitors (CGMs) produce readings every five minutes;
flash sensors return readings upon scanning, and meters upon fingerstick measurement.
The models represent sensor warm-up, data gaps, and compression artifacts. Scans and
fingerstick measurements are agent actions, making observation timing part of the control
policy. A reporting configuration $\rho$ additionally specifies the availability, timing,
and accuracy of reports about meals, activity, and other events.

\textbf{Action set.} Delivery devices determine the available treatment operations and
their constraints. The simulator includes \val{npumps} pump models, \val{npens} pen models,
and \val{ninsulins} insulin formulations. Each device configuration $d$ defines an action
set $\mathcal{A}_d$: pumps support device-specific subsets of boluses, temporary basal
rates, extended boluses, cancellation, and suspension or resumption; pens support discrete
injections. Insulin formulations determine the corresponding absorption and action profiles.
The device interface validates commands against dose increments, dose limits, bolus intervals,
and device state, and records rejected commands with their reasons. Occlusions and suspension alter
actual delivery, separating the requested intervention from the dose received.

\subsection{Instance Generation}
\label{sec:generator}

An instance specifies patient parameters, device and reporting configurations,
the event schedule, initial physiological and device states, pre-episode
history, evaluation horizon, and random seed. The generator assigns representative parameter values
or samples from configured ranges, then combines these specifications to produce
instances with different patients, devices, events, and observation conditions. Event
combinations, timing, and magnitude are configurable independently of the patient and
device parameters. Saving the resolved specification fixes the initialization, external
event schedule, and random seed for replay and comparisons between controllers.

\section{\benchname}
\label{sec:benchmark}

\benchname turns \simname into a fixed set of \val{nenvs} evaluation tasks. Each task is a
fixed patient--scenario configuration that specifies an individual, an event calendar, the
devices, the information available to the agent, the initial state, and the horizon, and each
is scored on the true glucose trajectory.

Existing medical benchmarks score answers, records, or plans for a single encounter or a
supplied history. \benchname instead scores multi-day control in which the agent's actions
shape what it observes next, and it computes every score programmatically from the true
trajectory (Table~\ref{tab:benchmark-comparison}).

\begin{table}[!htbp]
\centering
\caption{\textbf{Task scope and scoring in medical benchmarks.}}
\label{tab:benchmark-comparison}
\begingroup
\small
\setlength{\tabcolsep}{4pt}
\renewcommand{\arraystretch}{1.15}
\begin{tabularx}{\linewidth}{@{}>{\raggedright\arraybackslash}p{.36\linewidth}*{2}{>{\centering\arraybackslash}p{.13\linewidth}}>{\raggedright\arraybackslash}X@{}}
\toprule
Benchmark & Long-horizon control & Verifiable score & Evaluation tasks \\
\midrule
HealthBench~\citep{healthbench} & -- & $\circ$ & Medical QA \\
AgentClinic~\citep{agentclinic} & -- & $\circ$ & Diagnostic dialogue \\
MedAgentBench~\citep{medagentbench} & -- & $\checkmark$ & EHR tasks \\
LongMedBench~\citep{longmedbench} & $\circ$ & $\checkmark$ & Longitudinal clinical reasoning \\
DM-Bench~\citep{cardei2025dmbench} & $\circ$ & $\circ$ & Diabetes QA and planning \\
HealthLoopQA~\citep{niu2026healthloopqa} & $\circ$ & $\checkmark$ & Time-series QA \\
\midrule
\benchname (ours) & $\checkmark$ & $\checkmark$ & Closed-loop glucose control \\
\bottomrule
\end{tabularx}
\vspace{3pt}
\begin{minipage}{\linewidth}
\footnotesize
\textit{Long-horizon control:} $\checkmark$, multi-day interventions with action-dependent physiological feedback;
$\circ$, reasoning or planning over longitudinal records; --, encounter-level tasks.
\textit{Verifiable score:} $\checkmark$, fully programmatic scoring against references or rules;
$\circ$, includes model-based semantic grading.
\end{minipage}
\endgroup
\end{table}

\subsection{Task Design}
\label{sec:environments}

Within \simname's configurable environment space, \benchname defines
\val{ntemplates} scenario designs in \val{ngroups} groups. These cover meal timing and
absorption, residual insulin, exercise, changing insulin needs, pump under-delivery,
sensor errors, and illness recovery (Table~\ref{tab:scenarios}). Each design specifies
an event combination and its temporal relationships. Examples include a cancelled meal
after a pre-meal bolus followed by activity, or silent under-delivery with upward sensor
drift. We instantiate each design for four to eight simulated individuals, adapting event
timing and magnitude to each person, to obtain \val{nenvs} tasks. Appendix~\ref{app:main212-settings},
Table~\ref{tab:main212-events}, lists all designs by ID, including A01 and E03.

\begin{table}[!htbp]
\centering\small
\caption{\textbf{Scenario groups of \benchname.} Each design yields tasks for four to eight
individuals. Appendix~\ref{app:main212-settings} lists all \val{ntemplates} designs.}
\label{tab:scenarios}
\begin{tabularx}{\linewidth}{@{}lp{.25\linewidth}>{\raggedright\arraybackslash}Xcc@{}}
\toprule
 & Group & Example design & Designs & Tasks \\
\midrule
A & Meal timing and absorption & A03: a meal cancelled after pre-bolusing, followed by activity & 4 & 28 \\
B & Insulin from earlier doses & B01: repeated doses after a small meal overlap as absorption ends & 4 & 28 \\
C & Exercise & C04: breakfast and lunch skipped during prolonged aerobic activity & 4 & 24 \\
D & Changing insulin needs & D02: needs returning after sleep loss and the dawn phenomenon & 6 & 38 \\
E & Pump under-delivery & E04: silent under-delivery with progressively false-high sensor readings & 4 & 28 \\
F & Sensor errors & F06: missed meals and exercise causing lows masked by positive sensor bias & 6 & 36 \\
H & Illness and recovery & H03: insulin needs recover before appetite returns & 4 & 30 \\
\midrule
 & Total & & 32 & 212 \\
\bottomrule
\end{tabularx}
\end{table}

In the default setting, a task is a \val{mainhours}-hour episode with a Dana-i pump delivering
insulin aspart and real-time CGM with sensor error and delay, under information condition I2
described below. Because a configuration fixes the individual, the event calendar, and the
random seed, the same task can also be run with other delivery devices, information
conditions, or decision schedules, and a seven-day version of selected designs supports
multi-day evaluation with the same individual and personal record
(Section~\ref{sec:exp-evolve}). Section~\ref{sec:exp-ablations} uses these variants.

\noindent\textbf{Information conditions.} The reporting configuration $\rho$ takes three
levels. Under I2, the agent receives sensor glucose, the device state needed to operate the
pump, records of its own actions, and estimates derived from these records; meals and other
events are not reported. I1 adds the time and carbohydrate content of each meal. I0 further adds
free-text reports, such as stated meal plans and descriptions of upcoming events.

\subsection{Metrics}
\label{sec:scoring}

All metrics use the true glucose trajectory sampled every \val{gridminutes} minutes.
Let $g_1,\ldots,g_n$ denote these samples in mg/dL. We compute time in range
$\mathrm{TIR}=100n^{-1}\sum_{j=1}^{n}\mathbf{1}\{72.064\leq g_j\leq144.128\}$ and
time below range $\mathrm{TBR}=100n^{-1}\sum_{j=1}^{n}\mathbf{1}\{g_j<70\}$
as percentages. TIR uses 4--8~mmol/L (72.064--144.128~mg/dL) throughout the
main experiments and ablations. This range is tighter than the consensus target and is
specific to the benchmark, while the hypoglycemia threshold of 70~mg/dL follows the CGM
consensus~\citep{tirconsensus}.
The glucose standard deviation is
$\sigma_g=\sqrt{n^{-1}\sum_{j=1}^{n}(g_j-\bar g)^2}$, where
$\bar g=n^{-1}\sum_{j=1}^{n}g_j$.

\textbf{Overall score.} We adapt COGI, which combines time in range,
hypoglycemia, and glucose variability with weights of 0.50, 0.35, and
0.15~\citep{leelarathna2020cogi}. Our research variant \emph{COGI-4-8}
(abbreviated COGI below) replaces the published 70--180~mg/dL TIR band with
4--8~mmol/L, while retaining the hypoglycemia and population-SD components.
Let $c(x)=\min(1,\max(0,x))$. Its three subscores are
\begin{align*}
S_{\mathrm{TIR}} &= \mathrm{TIR},\\
S_{\mathrm{Hypo}} &= 100\,c\!\left(1-\frac{\mathrm{TBR}}{15}\right),\\
S_{\mathrm{GV}} &= 100\,c\!\left(\frac{108-\sigma_g}{90}\right).
\end{align*}
The hypoglycemia subscore decreases from 100 at zero TBR to zero at 15\% TBR.
The variability subscore decreases from 100 at a standard deviation of 18~mg/dL
to zero at 108~mg/dL. The overall score is
\[
S=0.50S_{\mathrm{TIR}}+0.35S_{\mathrm{Hypo}}+0.15S_{\mathrm{GV}}.
\]
COGI and all three subscores range from 0 to 100, with higher values indicating
better control. Results report their means across episodes with equal weight per task.

\section{Experiments}
\label{sec:results}
\label{sec:experiments}

\subsection{Setting}
\label{sec:exp-setting}

\noindent\textbf{Protocol.} We evaluate all methods on the \val{nenvs} tasks of
\benchname. Each 72-hour episode fixes the patient, initial state, event calendar, and
random seed, and uses a simulated Dana-i pump, insulin aspart, and real-time CGM under I2.
Meals and other events remain unreported. Each method acts through the same device
interface using only visible measurements and records; the environment continues to
evolve between decisions. The main evaluation adds no separate background HCL controller.
We compute COGI and its three subscores from 864 five-minute true-glucose samples per
episode, then average completed episodes with equal weight (Section~\ref{sec:scoring}).

\noindent\textbf{Baseline Setting.} We compare three controllers.
\emph{Basal--Bolus (BB)} follows a prescribed basal schedule and calculates meal and
correction boluses from reported carbohydrates, glucose, carbohydrate ratio, insulin
sensitivity, and insulin on board~\citep{holt2021consensus,fox2020rl}.
\emph{Event-Aware Basal--Bolus (BB+Rules)} augments this BB policy with predefined rules
for glucose trends, reported exercise and illness, and device problems~\citep{riddell2017exercise}.
\emph{OpenAPS} uses its open-source dosing core to compute temporary basal adjustments
and micro-boluses from glucose feedback. All three controllers use 15-minute decision
intervals and the same visible observations. Appendix~\ref{app:clinical} specifies the
formulas, rules, and OpenAPS implementation; the case studies also use OpenAPS as their reference.

\noindent\textbf{APC Setting.} We evaluate \apc with five language-model backends using a shared control objective, device-tool interface, and personal-record mechanism. The fixed decision interval is 60 minutes, with reasoning effort set to \texttt{high}.

\subsection{Main Results}
\label{sec:exp-main}

Table~\ref{tab:exp-main} reports the results. \apc with Claude Opus~5.5 has the highest
mean COGI, 73.54, followed by GPT~6.1 Sol (70.58), DeepSeek~4.1 Flash (68.61), Kimi~K3.1 Pro (67.15), and GLM~5.3
Flash (66.25). OpenAPS at a 15-minute interval achieves a COGI of 71.19.
\apc with Claude Opus~5.5 scores higher, while the other four models score lower.
Because more frequent model calls increase computational cost, we further examine
cooperation between a high-frequency background loop and lower-frequency agent decisions
in the CL-on/off experiments below.

\begin{table}[!htbp]
\centering\small
\caption{\textbf{Main evaluation.} APC rows average completed 72-hour episodes under I2 with a 60-minute decision interval. The three baselines use 15-minute decisions; APC uses \texttt{high} reasoning effort. COGI and its three unweighted subscores range from 0 to 100, higher being better.}
\label{tab:exp-main}
\setlength{\tabcolsep}{3pt}
\begin{tabular*}{\linewidth}{@{\extracolsep{\fill}}llcccc@{}}
\toprule
Controller / model & Interval & COGI & $S_{\mathrm{TIR}}$ & $S_{\mathrm{Hypo}}$ & $S_{\mathrm{GV}}$ \\
\midrule
\multicolumn{6}{l}{\emph{Baselines}} \\
BB~\citep{fox2020rl} & 15 min & 61.16 & 33.82 & 94.54 & 74.42 \\
BB+Rules~\citep{riddell2017exercise} & 15 min & 63.57 & 38.35 & 95.04 & 74.21 \\
OpenAPS~\citep{openaps} & 15 min & 71.19 & 51.62 & 96.45 & 77.48 \\
\midrule
\multicolumn{6}{l}{\emph{APC (backend model)}} \\
Claude Opus 5.5~\citep{anthropic2026opus55} & 60 min & \textbf{73.54} & 54.58 & 97.29 & 81.28 \\
GPT 6.1 Sol~\citep{openai2026gpt61sol} & 60 min & 70.58 & 49.07 & 97.93 & 78.42 \\
DeepSeek 4.1 Flash~\citep{deepseekai2026deepseekv41flash} & 60 min & 68.61 & 44.95 & 97.25 & 80.64 \\
Kimi K3.1 Pro~\citep{moonshot2026k31pro} & 60 min & 67.15 & 43.58 & 96.65 & 76.90 \\
GLM 5.3 Flash~\citep{glm5team2026glm5vibecodingagentic} & 60 min & 66.25 & 41.00 & 96.80 & 79.17 \\
\bottomrule
\end{tabular*}
\end{table}

\subsection{Ablations}
\label{sec:exp-ablations}

\subsubsection{Does Personal Experience Improve Control?}
\label{sec:exp-evolve}

We test whether personal records and end-of-day review improve control across days.
On six configurations over seven days, we compare four conditions:
\textbf{daily record reset}, which passes no personal record to the next day;
\textbf{cross-day record only}, which retains personal records for the next day without
additional end-of-day review; \textbf{record + review-generated plan}, which retains
records and reviews the day's treatment responses to generate a plan for subsequent
control; and \textbf{record + plan + execution feedback}, which additionally incorporates
feedback on the execution of previous plans into the review. All conditions reset the
conversation daily while maintaining continuous physiology and fixed model weights.

\begin{figure}[!htbp]
\centering
\includegraphics[width=\linewidth]{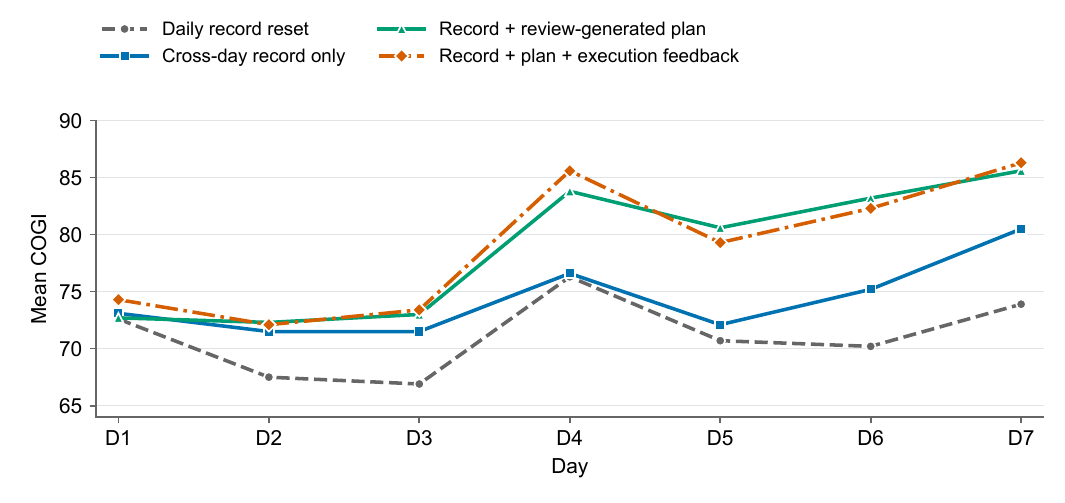}
\caption{\textbf{Daily scores under the four record and review conditions.} Each point is the mean COGI across six configurations in the corresponding 24-hour window, matching Table~\ref{tab:exp-days}(a). Lines connect daily means without smoothing. The four conditions are defined in the text.}
\label{fig:daily-cogi}
\end{figure}

Retaining records raises whole-week COGI from 71.1 to
74.4; adding review and planning raises it to 78.7 (Table~\ref{tab:exp-days}). Adding
execution feedback raises it by another 0.3 points. Figure~\ref{fig:daily-cogi} shows the
daily scores under all four conditions. Gains mainly come from time in range, and all
conditions retain the maximum hypoglycemia subscore. Over this week, records and review
support improvement without weight updates; reviewing past responses helps the agent
use accumulated experience in subsequent control.

\begin{table}[!htbp]
\centering\small
\caption{\textbf{Seven-day record and review ablation.} Six held-out configurations;
Claude Opus~5.5, Dana-i, real-time CGM, I2, fixed weights, one run per condition and configuration.
Historical runs use an earlier prompt; we recompute COGI from saved glucose.
Review adds model calls, with the treatment-decision budget unchanged. Entries average the six configurations.}
\label{tab:exp-days}
\setlength{\tabcolsep}{4pt}
\renewcommand{\arraystretch}{1.15}
\textbf{(a) Daily COGI}\\[3pt]
\begin{tabularx}{\linewidth}{@{}X*{7}{c}@{}}
\toprule
Condition & D1 & D2 & D3 & D4 & D5 & D6 & D7\\
\midrule
Daily record reset & 72.6 & 67.5 & 66.9 & 76.3 & 70.7 & 70.2 & 73.9\\
Cross-day record only & 73.1 & 71.5 & 71.5 & 76.6 & 72.1 & 75.2 & 80.5\\
Record + review-generated plan & 72.7 & 72.3 & 73.0 & 83.8 & 80.6 & 83.2 & 85.6\\
Record + plan + execution feedback & 74.3 & 72.1 & 73.4 & 85.6 & 79.3 & 82.3 & 86.3\\
\bottomrule
\end{tabularx}
\par\medskip
\textbf{(b) Whole-week COGI and subscores}\\[3pt]
\begin{tabularx}{\linewidth}{@{}Xcccc@{}}
\toprule
Condition & COGI & $S_{\mathrm{TIR}}$ & $S_{\mathrm{Hypo}}$ & $S_{\mathrm{GV}}$\\
\midrule
Daily record reset & 71.1 & 43.3 & 100.0 & 96.2\\
Cross-day record only & 74.4 & 49.4 & 100.0 & 97.9\\
Record + review-generated plan & 78.7 & 58.0 & 100.0 & 97.9\\
Record + plan + execution feedback & 79.0 & 58.6 & 100.0 & 98.1\\
\bottomrule
\end{tabularx}
\par\medskip
\begin{minipage}{\linewidth}\footnotesize
Whole-week scores use pooled seven-day samples; daily scores use separate 24-hour windows.
\end{minipage}
\end{table}

\subsubsection{How Do Delivery Devices Affect Control?}
\label{sec:exp-devices}

\begin{figure}[!htbp]
\centering
\includegraphics[width=\linewidth]{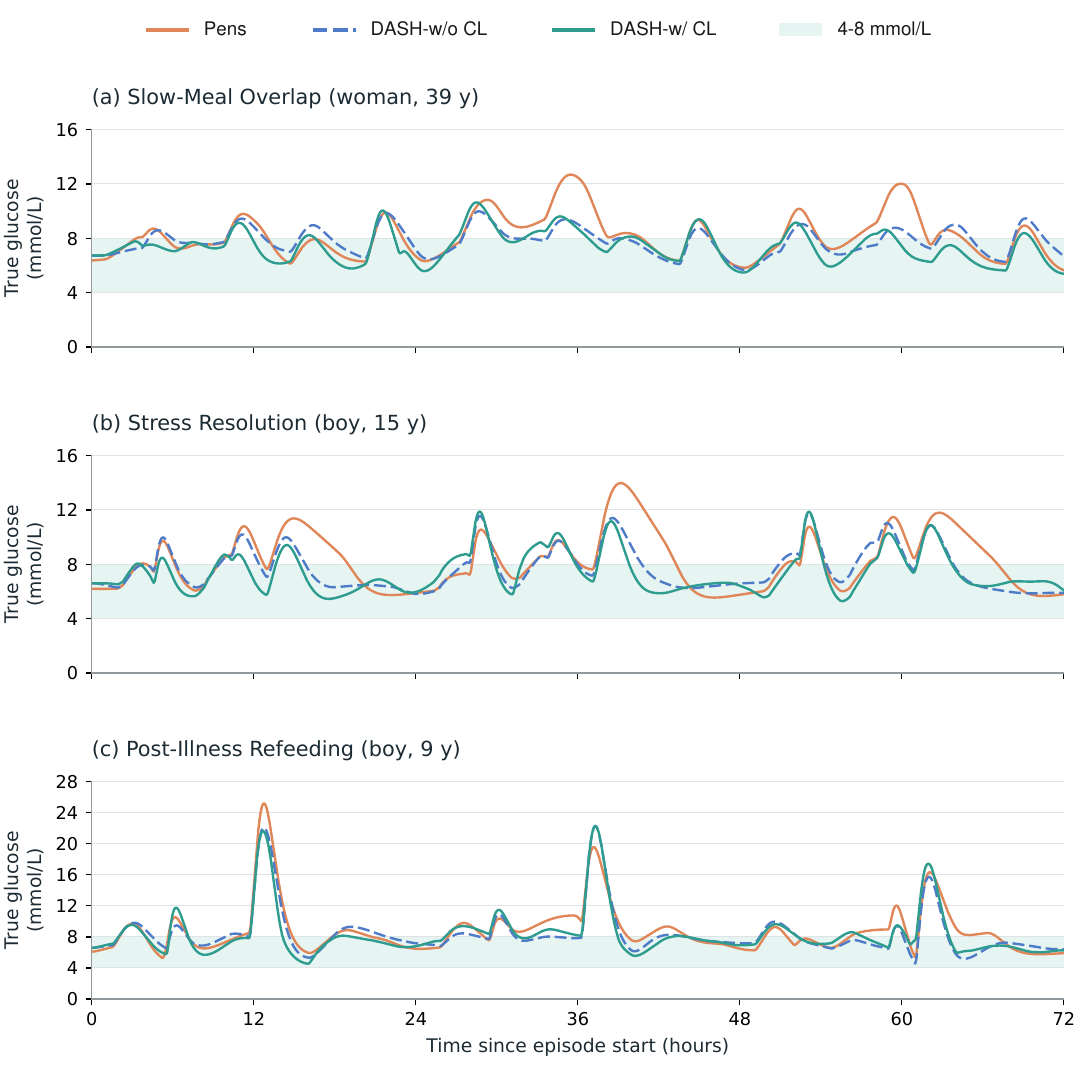}
\caption{\textbf{Three hardware trajectories within each scenario.}
Each panel overlays the complete 72-hour true-glucose trajectories for the
same patient and exogenous event sequence: pens, DASH-w/o CL (without background
closed-loop control), and DASH-w/ CL (with background closed-loop control).
The pale teal horizontal band is 4--8~mmol/L, and time is relative to each episode start.}
\label{fig:hardware-trajectories}
\end{figure}

We compare pens and DASH with CL off or on in Slow-Meal Overlap, Stress Resolution,
and Post-Illness Refeeding, testing both delivery options and cooperation with a
background controller. Each configuration uses the same agent for a 72-hour run in
each case. DASH-w/o CL scores approximately 6--8 points above pens, mainly through time
in range and variability (Table~\ref{tab:exp-devices}). Turning CL on further raises
COGI by 4.63 and 2.53 points in the first two cases and gives a similar score in the
third, without increasing the agent's decision frequency. The background HCL supplies
frequent basal-rate adjustments, while the agent integrates observations and personal
history to choose doses, carbohydrate intake, and device actions. The trajectories show
how adjustable infusion follows prolonged absorption and changing insulin needs
(Figure~\ref{fig:hardware-trajectories}). These cases show that delivery configurations
shape control performance and support the potential of high-frequency closed-loop
control combined with lower-frequency agent decisions.

We additionally evaluate standalone closed-loop control without an agent every five minutes,
using the same DASH/Fiasp configuration, patients, seeds, events, and CGM conditions,
with rule-based sensor maintenance. Agent + closed-loop control exceeds this reference
by 13.42, 3.63, and 6.84 COGI points in the three cases, respectively, mainly through
more time in range.

\begin{table}[!htbp]
\centering\small
\caption{\textbf{Delivery devices and background closed-loop control.} Each configuration runs for 72 hours per case; APC uses Claude Opus~5.5. Bold marks the highest COGI per case.}
\label{tab:exp-devices}
\setlength{\tabcolsep}{3pt}
\renewcommand{\arraystretch}{1.12}
\begin{tabular*}{\linewidth}{@{\extracolsep{\fill}}l*{9}{r}@{}}
\toprule
Configuration & \multicolumn{3}{c}{Slow-Meal Overlap} & \multicolumn{3}{c}{Stress Resolution} & \multicolumn{3}{c}{Post-Illness Refeeding} \\
 & \multicolumn{3}{c}{\footnotesize Woman, 39 y} & \multicolumn{3}{c}{\footnotesize Boy, 15 y} & \multicolumn{3}{c}{\footnotesize Boy, 9 y} \\
\cmidrule(lr){2-4}\cmidrule(lr){5-7}\cmidrule(l){8-10}
 & COGI & $S_{\mathrm{TIR}}$ & $S_{\mathrm{GV}}$ & COGI & $S_{\mathrm{TIR}}$ & $S_{\mathrm{GV}}$ & COGI & $S_{\mathrm{TIR}}$ & $S_{\mathrm{GV}}$ \\
\midrule
APC + Pens & 74.61 & 52.55 & 88.90 & 72.33 & 51.04 & 78.71 & 67.10 & 46.76 & 58.11 \\
APC + DASH-w/o CL & 80.56 & 61.11 & 100.00 & 80.22 & 63.66 & 89.29 & \textbf{74.67} & 60.88 & 61.57 \\
APC + DASH-w/ CL & \textbf{85.19} & 71.18 & 97.33 & \textbf{82.75} & 68.52 & 89.93 & 74.51 & 60.88 & 60.49 \\
\addlinespace[3pt]
DASH-w/ CL only & 71.77 & 43.87 & 98.93 & 79.12 & 61.11 & 90.41 & 67.67 & 48.84 & 55.02 \\
\bottomrule
\end{tabular*}
\par\smallskip
\begin{minipage}{\linewidth}\footnotesize
All runs have $S_{\mathrm{Hypo}}=100$ and zero time below 70~mg/dL; $S_{\mathrm{TIR}}$ uses 4--8~mmol/L. Pens use aspart and degludec, with at most five injections per day; DASH uses Fiasp. CL denotes the default background HCL in our DASH configuration. APC retains the main decision schedule with CL off or on; CL alone runs every five minutes with rule-based sensor maintenance. See Appendix~\ref{app:clinical} for implementation.
\end{minipage}
\end{table}

\subsubsection{How Do Information and Decision Frequency Affect Control?}
\label{sec:exp-frequency}
\label{sec:exp-observation}
\label{sec:exp-rare}

\begingroup
\clubpenalty=10000
We test the effects of information and decision frequency in Infusion-Site Leakage and
Mixed-Intensity Exercise. Each scenario crosses three information levels with fixed
intervals of 15, 30, 60, and 120 minutes, giving 24 runs of 72 hours each with the same
extra-decision allowance. At a fixed 60-minute interval, moving from I2 to I0 raises COGI
by 43.2 points on leakage and 15.7 on exercise (Table~\ref{tab:exp-frequency}). Under I2,
shortening the interval from 120 to 15 minutes adds 10.5 and 3.7 points, respectively.
Averaged over both scenarios and all information levels, input-token use at 15 minutes
is about 4.2 times that at 60 minutes. In these two cases, added context yields larger
gains than increasing frequency alone. Frequent decisions partly compensate for missing
information but require more computation. Together with the CL-on/off results and the
interval sweep for OpenAPS (Appendix~\ref{app:clinical}), this supports using frequent
background control alongside an agent that makes informed, less frequent decisions.
\par
\endgroup

\begin{table}[!htbp]
\centering\small
\caption{\textbf{Information and decision-frequency comparison.} Claude Opus~5.5 under the
main protocol; one 72-hour run per cell, 24 in total. I2 provides device observations and
own records; I1 adds meal times and carbohydrate amounts; I0 adds free-text context.
All arms share the same patient, seed, initial state, and events within each scenario,
with at most six extra decisions per local day. Entries are COGI-4-8 totals.}
\label{tab:exp-frequency}
\setlength{\tabcolsep}{10pt}
\renewcommand{\arraystretch}{1.12}
\begin{adjustbox}{max width=\linewidth}
\begin{tabular*}{\linewidth}{@{\extracolsep{\fill}}llrrrr@{}}
\toprule
Scenario / patient & Information & 15 min & 30 min & 60 min & 120 min \\
\midrule
\multirow{3}{*}{\shortstack[l]{Infusion-Site Leakage\\Man, 39 y}} & I0 & 97.80 & 98.96 & 98.38 & 99.71 \\
 & I1 & 66.70 & 69.57 & 60.07 & 56.00 \\
 & I2 & 62.10 & 59.94 & 55.19 & 51.59 \\
\midrule
\multirow{3}{*}{\shortstack[l]{Mixed-Intensity Exercise\\Man, 39 y}} & I0 & 98.84 & 98.67 & 97.86 & 94.39 \\
 & I1 & 98.78 & 96.18 & 96.29 & 92.55 \\
 & I2 & 85.27 & 85.41 & 82.12 & 81.59 \\
\bottomrule
\end{tabular*}
\end{adjustbox}
\end{table}

\subsection{Case Study}
\label{sec:exp-case-study}

We examine three episodes of \apc with Claude Opus~5.5 from the main evaluation: two
where it scores above OpenAPS and one where it scores below.
Table~\ref{tab:exp-cases} gives their scores. In C04 the advantage comes mainly from reducing
hypoglycemia; in F06 the agent also improves time in range when sensor bias masks low
glucose. In E04 both controllers avoid hypoglycemia completely, and the agent falls behind
on time in range and variability. These are illustrative cases, not estimates of average effects.

\begin{table}[!htbp]
\centering\small
\caption{\textbf{Whole-episode outcomes for the three case studies.} Claude Opus~5.5
and OpenAPS in the same world under I2; each row is one 72-hour episode.
COGI and its three unweighted subscores range from 0 to 100, higher being better.
Bold indicates the better value within each case, including ties.}
\label{tab:exp-cases}
\setlength{\tabcolsep}{5pt}
\begin{tabular*}{\linewidth}{@{\extracolsep{\fill}}llcccc@{}}
\toprule
Case & Controller & COGI & $S_{\mathrm{TIR}}$ & $S_{\mathrm{Hypo}}$ & $S_{\mathrm{GV}}$ \\
\midrule
\multirow{2}{*}{C04, adult female} & Claude Opus 5.5 & \textbf{85.06} & 84.14 & \textbf{83.02} & 92.85 \\
 & OpenAPS & 77.73 & \textbf{85.42} & 59.10 & \textbf{95.57} \\
\midrule
\multirow{2}{*}{F06, child female} & Claude Opus 5.5 & \textbf{72.75} & \textbf{68.52} & \textbf{76.85} & \textbf{77.26} \\
 & OpenAPS & 48.52 & 57.18 & 26.70 & 70.61 \\
\midrule
\multirow{2}{*}{E04, adolescent female} & Claude Opus 5.5 & 47.37 & 10.65 & \textbf{100.00} & 47.00 \\
 & OpenAPS & \textbf{66.20} & \textbf{44.56} & \textbf{100.00} & \textbf{59.48} \\
\bottomrule
\end{tabular*}
\end{table}

\begin{figure}[!htbp]
\centering
\includegraphics[width=\linewidth]{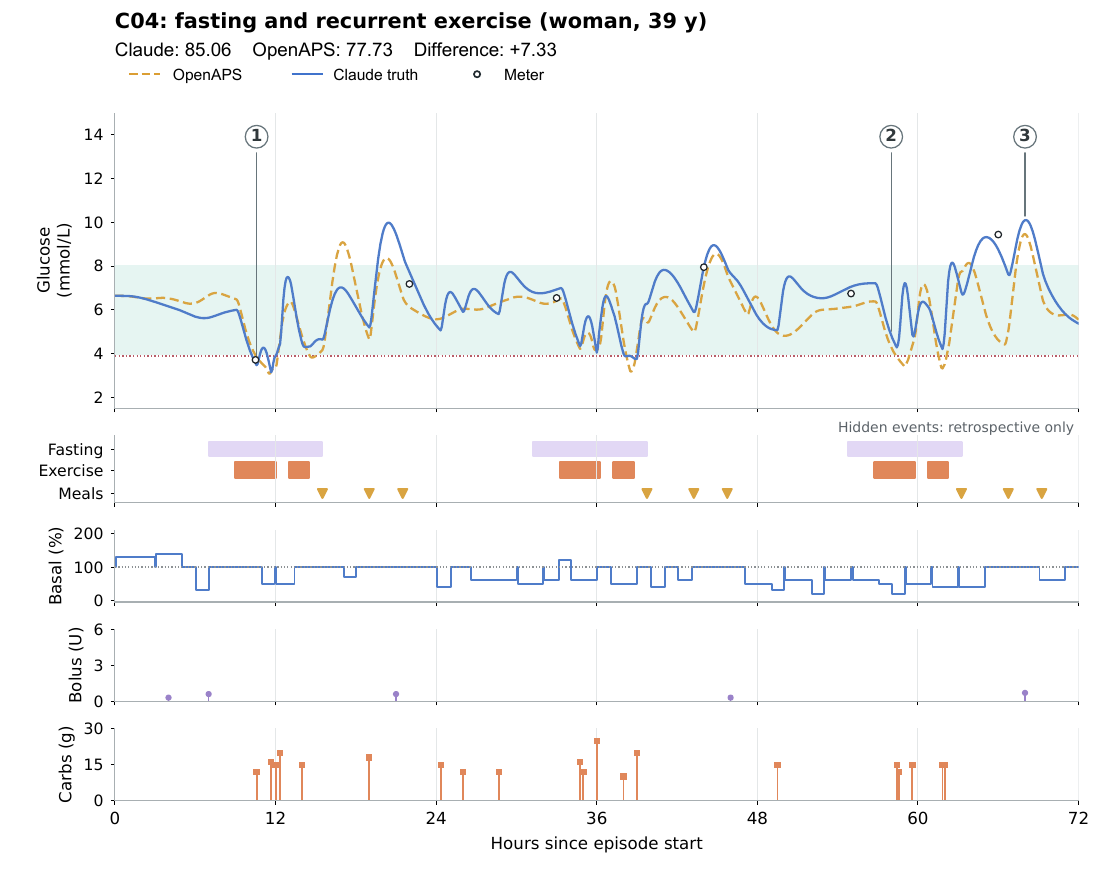}
\caption{\textbf{Stronger control: C04, adult female.} Solid blue shows true
glucose under agent control; open circles mark meter checks;
dashed gold shows OpenAPS true glucose in the
same world; and the pale teal band marks 4--8 mmol/L. The event lane shows hidden
fasting, exercise, and meals. Lower lanes show realized basal delivery relative
to scheduled basal, boluses, and carbohydrate rescue. Numbered markers identify
the decisions discussed in the text.}
\label{fig:case-good}
\end{figure}

\paragraph{Responding to an unreported cause of falling glucose.}
In C04, a 39-year-old woman skips breakfast and lunch and exercises repeatedly for long periods
before a late meal, so hypoglycemia is a constant risk. \apc scores 85.06 against 77.73 for
OpenAPS ($+7.33$). Its time in 4--8~mmol/L is 84.14\%, slightly below OpenAPS's 85.42\%, but
its time below 70~mg/dL is 2.55\%, compared with 6.13\% for OpenAPS.
Figure~\ref{fig:case-good} marks three decisions. At marker~1 (10.58~h), a meter check
confirms a low of 67~mg/dL and the agent gives 12~g of fast carbohydrate. As glucose keeps
falling with little insulin recorded as active, the agent's rationale raises the possibility of
unreported exercise, and it responds with carbohydrate and temporary basal
reductions that it can later reverse. At marker~2 (58~h), its record of low glucose in the late
morning on the two previous days leads it to lower the temporary basal rate to 20\% early, and
it adds carbohydrate at 58.42 and 58.58~h as glucose continues to drop. At marker~3 (68~h),
having seen a strong fall after an earlier correction, it gives only 0.7~U, and when glucose
falls faster than it predicted, it lowers basal delivery again. In this episode the agent checks
the sensor against the meter, revises its explanation when the response does not match, and
carries lessons from earlier days into later doses.

\begin{figure}[!htbp]
\centering
\includegraphics[width=\linewidth]{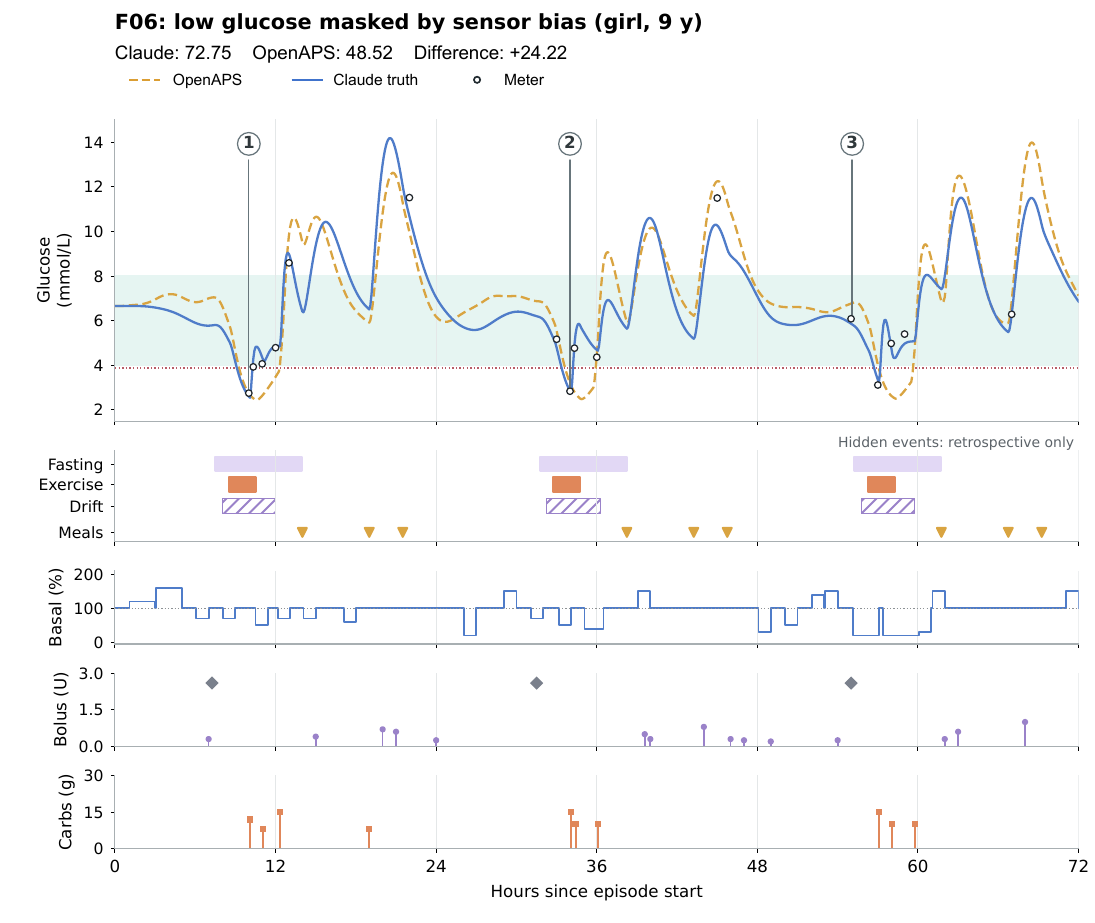}
\caption{\textbf{Stronger control under sensor bias: F06, child female.} Curve and meter
conventions match Figure~\ref{fig:case-good}, with OpenAPS as the comparator.
The event lane shows hidden fasting, exercise, positive sensor drift, and meals.
Lower lanes show the agent's basal changes, boluses, and carbohydrate rescue; gray diamonds
mark preset meal boluses from the event calendar, not agent decisions.
Markers identify meter verification on the first two mornings and an earlier basal reduction
on the third. Hidden events are retrospective annotations, not agent inputs.}
\label{fig:case-sensor}
\end{figure}

\paragraph{Detecting low glucose hidden by sensor bias.}
In F06, a 9-year-old girl misses breakfast after a meal bolus and then exercises while a
positive sensor bias masks the fall in true glucose. \apc scores 72.75 against 48.52 for
OpenAPS ($+24.22$). Time in 4--8~mmol/L is 68.52\% versus 57.18\%, and time below
70~mg/dL is 3.47\% versus 11.00\%. Figure~\ref{fig:case-sensor} highlights three decisions.
At marker~1 (10~h), calibration gives a meter reading of 50~mg/dL when CGM had read about
94; the agent gives 12~g of fast carbohydrate five minutes later and checks again.
At marker~2 (34~h), it explicitly recalls the earlier positive bias and obtains a meter
reading of 51 despite CGM near 91. It gives 15~g and confirms a rise to 86~mg/dL fifteen
minutes later. At marker~3 (55.08~h), its rationale cites the lows on both earlier mornings
and starts a two-hour temporary basal rate at 20\% before the recurring decline.
Lows still occur and require further carbohydrate rescue. This case shows how the agent can
combine repeated meter checks with saved experience to adjust earlier and reduce hypoglycemia.

\begin{figure}[!htbp]
\centering
\includegraphics[width=\linewidth]{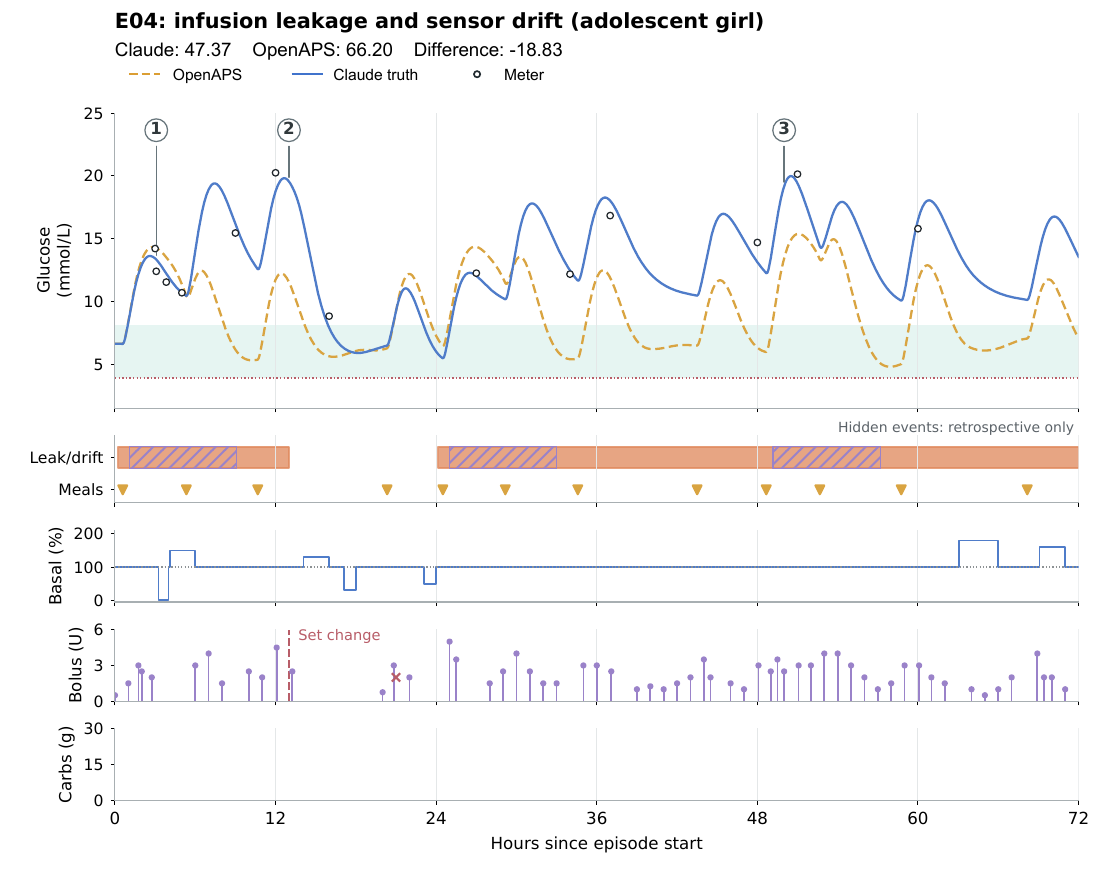}
\caption{\textbf{Weaker control: E04, adolescent female.} Curve and action
conventions match Figure~\ref{fig:case-good}. Orange shading in the event lane
marks evaluator-confirmed reduced tissue delivery; hatching marks programmed
sensor-drift intervals, and triangles mark hidden meals. The vertical dashed
line in the bolus lane is the sole infusion-set change; the red cross marks
the rejected bolus request. The empty carbohydrate lane indicates that no
carbohydrate rescue was issued. The repair and later recurrence can be verified
from effective delivery, whereas the agent sees only measurements and device
and action records.}
\label{fig:case-bad}
\end{figure}

\paragraph{Losing track of a recurring delivery fault.}
In E04, an adolescent girl has recurring infusion-site leakage and intermittent positive sensor
drift. \apc scores 47.37 against 66.20 for OpenAPS ($-18.83$), with 10.65\% of the time in
4--8~mmol/L and no true glucose below 70~mg/dL, so the deficit comes from persistent
hyperglycemia and high variability. Near marker~1 in Figure~\ref{fig:case-bad}, the agent uses
meter checks and calibration to detect the sensor disagreement, but it repeatedly attributes the
weak response to insulin to unreported food or changing insulin needs. It changes the infusion
set at 13~h (marker~2), about 12.75~h after the leak began, and the saved simulator state
confirms that effective delivery then recovers from 15\% to 100\%. When the leak returns near
24.17~h and 48.33~h, however, it does not change the set again. At marker~3 (50~h), with CGM at
its upper display limit, it gives another 2.5~U. Over the episode it requests 60 boluses, of
which 59 are delivered and one is rejected by the minimum-interval rule, and it changes the set
only once. The agent found the fault once but did not test that explanation again when the same
symptoms returned, and it kept adding insulin instead of looking for the cause.

\section{Discussion}
\label{sec:discussion}

\subsection{Decision Making under Incomplete Information}
\label{sec:discussion-information}

The information ablation shows that agents do not yet combine measurements, event context,
and personal history effectively. Added context improves control more than increasing
decision frequency alone in the tested cases, and frequent decisions only partly compensate
for what the agent cannot see. The E04 episode shows the same weakness within a single run:
the agent identified a delivery fault once, then kept adding insulin when the same symptoms
returned instead of testing that explanation again. Explicit uncertainty estimates and
targeted measurement requests would let an agent separate changing insulin needs from
delivery or sensor problems, and personal records can carry those inferences into later
reassessment.

\subsection{Safety}
\label{sec:discussion-safety}

Insulin continues to act after delivery, making dose history and reassessment timing
central to physiological control \citep{holt2021consensus}. An agent that doses insulin
therefore needs dose limits, insulin-on-board estimation, and low-glucose monitoring.
The CL-on/off cases also argue for combining frequent background regulation with less
frequent agent decisions. Separate hypoglycemia scores and action records let these
designs be assessed alongside overall control performance.

\section{Related Work}
\label{sec:related-work}

\paragraph{Medical AI and benchmarks.}
MultiMedQA and HealthBench assess clinical knowledge and health-related responses,
while AMIE studies diagnostic dialogue and longitudinal disease management
\citep{multimedqa,healthbench,amie2025diagnosis,amie2026management}. AgentMD and TxAgent
connect medical reasoning to clinical calculators and biomedical evidence
\citep{jin2025agentmd,gao2025txagent}. AgentClinic, MedAgentBench, LongMedBench, and
HealthAgentBench evaluate agents through simulated consultations, electronic health
records, longitudinal data, and clinical tool workflows
\citep{agentclinic,medagentbench,longmedbench,liu2026healthagentbench}.

\paragraph{Long-horizon agents.}
ReAct interleaves reasoning with environment actions; WebArena, OSWorld, and SWE-agent
study extended tool use in web, desktop, and software tasks
\citep{react2023,zhou2023webarena,xie2024osworld,sweagent2024}. SayCan and Code as Policies
connect language-based planning to robot skills and control interfaces
\citep{ahn2022saycan,liang2022codeaspolicies}. Reflexion, ExpeL, and Voyager retain feedback,
experience, or executable skills to improve later decisions without updating model weights
\citep{reflexion2023,expel2024,voyager2023}.

\paragraph{Type~1 diabetes and glucose control.}
CGM-Agent analyzes personal glucose records; DM-Bench and HealthLoopQA evaluate decisions
and temporal reasoning over supplied histories
\citep{cui2026cgmagent,cardei2025dmbench,niu2026healthloopqa}. DIETS and meal-planning agents
optimize candidate interventions using glucose forecasts
\citep{zeng2024diets,huang2026mealagent}. UVA/Padova and simglucose support virtual-patient
evaluation, while automated insulin delivery and reinforcement-learning controllers
adjust treatment from glucose feedback
\citep{visentin2018uva,simglucose,brown2019closedloop,sherr2022aid,hettiarachchi2024g2p2c,khamesian2026guide}.
LLM-T1D distills a reinforcement-learning policy into a language model for insulin-rate
control and evaluates it in randomized 24-hour simulations \citep{sarkar2026llmt1d}.

\section{Limitations and Future Work}
\label{sec:future-work}

\paragraph{Simulator fidelity.}
The gap between \simname and real physiology bounds what these results say about use with
patients. We plan to reconstruct
patient-specific environments from real CGM and device records, which can inform
physiology, sensor errors, and event patterns, with held-out records assessing the
reconstruction. An agent would then build its record and strategy inside a person's
reconstructed environment before assisting that person, and each new observation would
refine both the environment and the agent.

\paragraph{Reinforcement learning.}
\apc adapts through its personal record while its weights stay fixed. \simname also
supports training on physiological outcomes directly. Reinforcement learning can update
the policy for treatment, measurement, and reassessment timing, and combining it with the
record would let the agent acquire general skills and person-specific knowledge together.

\paragraph{Beyond type~1 diabetes.}
\apc applies wherever devices can measure and act on a physiological system and its outcome
can be scored. Candidate domains include other conditions managed with drug-delivery devices
and intensive-care treatment, where learned treatment policies have already been studied
\citep{komorowski2018aiclinician}. Each new domain needs its own devices, environment, and
score, and we hope to build them together with the community.

\section{Conclusion}
\label{sec:conclusion}

We introduce \fwname, comprising the \apc framework, the \simname simulator, and the \benchname benchmark.
\apc links agent decisions to medical-device actions and applies this control loop to glucose management in type~1 diabetes.
\simname provides virtual testing across patients, events, devices, and observation conditions.
\benchname turns this environment into \val{nenvs} tasks for evaluating the physiological outcomes of agent decisions.
Together, these components establish a reproducible research setting for agentic physiological control.
We hope \fwname provides a foundation for agents to participate in human physiological control and supports extending this research to other physiological systems.

\section*{Acknowledgments}
We thank Einsia AI for sponsoring this work. We also thank the volunteers from the Sweetie Online Community for their professional advice and testing.

\bibliographystyle{plainnat}
\bibliography{refs}

\appendix

\section{Scenario Designs}
\label{app:main212-settings}

Table~\ref{tab:main212-events} lists all 32 scenario designs and their 212 evaluation tasks. IDs use a letter and two digits: A1 and E3 denote A01 and E03, respectively. Event timing and magnitude vary by patient configuration.

{\small
\setlength{\tabcolsep}{4pt}
\begin{longtable}{@{}p{0.08\linewidth}p{0.81\linewidth}r@{}}
\caption{\textbf{\benchname designs and their event combinations.} Each design yields evaluation tasks for different patients.}
\label{tab:main212-events}\\
\toprule
ID & Event combination & Tasks \\
\midrule
\endfirsthead
\toprule
ID & Event combination & Tasks \\
\midrule
\endhead
\bottomrule
\endfoot
A01 & Repeated small intakes overlapping a slowly absorbed main meal & 6 \\
A02 & Two slow-meal tails overlapping subsequent intake & 8 \\
A03 & A meal cancelled after pre-bolusing, followed by activity & 8 \\
A04 & Breakfast cancelled after a meal bolus, followed by prolonged fasting & 6 \\
B01 & Repeated dosing after a small meal; absorption ends while insulin effects overlap & 6 \\
B02 & Activity and small meals after earlier stacked doses & 8 \\
B03 & Remaining extended delivery after a cancelled meal, with delayed replacement intake & 6 \\
B04 & Rescue carbohydrate still absorbing while the main meal is delayed & 8 \\
C01 & Two activity bouts interleaved with replenishment and meals & 8 \\
C02 & A transient increase in hepatic glucose output followed by aerobic activity & 6 \\
C03 & Delayed effects of evening activity and alcohol & 4 \\
C04 & Breakfast and lunch skipped during prolonged aerobic activity & 6 \\
D01 & Two short stress episodes resolving before subsequent activity & 6 \\
D02 & Insulin needs returning after sleep loss and the dawn phenomenon & 8 \\
D03 & Dawn-related needs receding during slow breakfast absorption and later fast intake & 6 \\
D04 & Sleep-loss-related needs recovering during lunch absorption and later intake & 8 \\
D05 & Two post-meal stress episodes resolving during slow and fast intake & 6 \\
D06 & A slow overnight meal tail overlapping delayed alcohol effects and next-morning intake & 4 \\
E01 & Silent under-delivery overlapping multiple meals & 8 \\
E02 & Three intermittent under-delivery episodes with irregular meals & 6 \\
E03 & Active repair of under-delivery followed by activity & 8 \\
E04 & Silent under-delivery combined with progressively false-high sensor readings & 6 \\
F01 & An activity-related glucose fall combined with false-high sensor readings & 6 \\
F02 & Real post-meal glucose input combined with false-low sensor readings & 6 \\
F03 & Meals and activity around a sensor-signal gap & 6 \\
F04 & Illness recovery combined with incorrect calibration and sensor drift & 6 \\
F05 & Calibration expiry with residual slow-meal absorption and recent dosing & 6 \\
F06 & Missed meals and exercise causing low glucose, masked by positive sensor bias & 6 \\
H01 & Repeated meals during illness; slow absorption persists into recovery & 6 \\
H02 & Reduced intake during illness with pre-episode slow absorption and recent treatment & 8 \\
H03 & Insulin needs recover before appetite; slow lunch absorption precedes fast intake & 8 \\
H04 & Illness with under-delivery requiring repair, followed by recovery of insulin needs & 8 \\
\end{longtable}
}

\section{Simulator Specification}
\label{app:interface}
\label{app:library}
\label{app:hardware-actions}

This section specifies monitoring interfaces and delivery-device constraints.

\subsection{Observation Interface}
Each decision receives device readings and state, visible delivery and event records,
and insulin- and carbohydrate-on-board estimates. Real-time CGM supplies readings
automatically; scan sensors and meters provide readings through scan or fingerstick
actions. The information condition determines which event reports are visible, while
sensor state determines reading availability and freshness.
Table~\ref{tab:hardware-sensing} lists the monitoring interfaces.

\begin{table}[htbp]
\centering\footnotesize
\setlength{\tabcolsep}{3pt}
\renewcommand{\arraystretch}{1.18}
\caption{\textbf{Sensing-device capability matrix.} Each column describes the device
itself; a separately configured meter adds fingerstick measurement to any sensor
combination. $\checkmark^{m}$ requires that meter; $\checkmark^{a}$ requires the
optional alarm to be configured.}
\label{tab:hardware-sensing}
\begin{tabularx}{\linewidth}{@{}p{0.25\linewidth}*{7}{>{\centering\arraybackslash}X}@{}}
\toprule
Function / parameter & Meter & \shortstack{Dexcom\\G5} & \shortstack{Dexcom\\G6/G7} & \shortstack{Guardian\\3} & \shortstack{Libre 2/3\\real-time} & \shortstack{Libre 1\\scan} & \shortstack{Libre 2\\scan}\\
\midrule
\code{MEASURE\_BG} from this device & $\checkmark$ & $\times$ & $\times$ & $\times$ & $\times$ & $\times$ & $\times$\\
\code{SCAN} & $\times$ & $\times$ & $\times$ & $\times$ & $\times$ & $\checkmark$ & $\checkmark$\\
\code{CALIBRATE\_CGM} & $\times$ & $\checkmark^{m}$ & $\times$ & $\checkmark^{m}$ & $\times$ & $\times$ & $\times$\\
High/low-only alarm before scan & $\times$ & $\times$ & $\times$ & $\times$ & $\times$ & $\times$ & $\checkmark^{a}$\\
\midrule
Observation mode & Point & Real-time & Real-time & Real-time & Real-time & Scan & Scan\\
Simulated sensor interval (min) & Per check & 5 & 5 & 5 & 5 & 5 & 5\\
\bottomrule
\end{tabularx}
\end{table}

\begin{samepage}
\subsection{Action Sets}
The delivery configuration specifies available operations, dose units, bounds, and steps.
Pump operations include boluses, temporary basal rates, suspension, and infusion-set
changes; pens support discrete injections. The device interface validates each command,
uses device state to determine actual delivery, and returns the execution result.
Waiting and carbohydrate intake are common to all configurations.
Table~\ref{tab:hardware-delivery} summarizes the simulated device constraints, drawing on
manufacturer documentation and AndroidAPS drivers~\citep{androidaps}.

\end{samepage}
\begin{table}[htbp]
\centering\footnotesize
\setlength{\tabcolsep}{3pt}
\renewcommand{\arraystretch}{1.18}
\caption{\textbf{Delivery-device capability matrix.} $\checkmark$: the function is
exposed for this model; $\times$: not exposed.}
\label{tab:hardware-delivery}
\begin{tabularx}{\linewidth}{@{}p{0.26\linewidth}*{6}{>{\centering\arraybackslash}X}@{}}
\toprule
Function / parameter & \shortstack{Pens\\(MDI)} & Combo & DASH & Dana-i & \shortstack{Medtrum\\Nano} & Insight\\
\midrule
Ordinary bolus & $\checkmark$ & $\checkmark$ & $\checkmark$ & $\checkmark$ & $\checkmark$ & $\checkmark$\\
Long-acting insulin injection & $\checkmark$ & $\times$ & $\times$ & $\times$ & $\times$ & $\times$\\
Set / cancel temporary basal & $\times$ & $\checkmark$ & $\checkmark$ & $\checkmark$ & $\checkmark$ & $\checkmark$\\
Suspend / resume pump & $\times$ & $\checkmark$ & $\checkmark$ & $\checkmark$ & $\checkmark$ & $\checkmark$\\
Infusion-site / pod change & $\times$ & $\checkmark$ & $\checkmark$ & $\checkmark$ & $\checkmark$ & $\checkmark$\\
Independent ordinary-bolus cancellation & $\times$ & $\times$ & $\checkmark$ & $\checkmark$ & $\checkmark$ & $\checkmark$\\
Set / cancel extended bolus & $\times$ & $\times$ & $\times$ & $\checkmark$ & $\times$ & $\checkmark$\\
\midrule
Bolus increment (U) & Pen-specific & 0.1 & 0.05 & 0.05 & 0.05 & 0.01\\
Bolus speed (U/min) & Immediate & 12 & 1.5 & 5 & 1.5 & 6\\
Native temporary-basal type & -- & Percent & Absolute & Percent & Absolute & Percent\\
Temporary-basal increment & -- & 10\% & 0.05\,U/h & 10\% & 0.05\,U/h & 10\%\\
Simulated temporary-basal maximum & -- & 500\% & 30\,U/h & 200\% & 25\,U/h & 250\%\\
Temporary-basal duration grid (min) & -- & 15 & 30 & 15 & 30 & 15\\
\bottomrule
\end{tabularx}
\end{table}

\section{Evaluation Details}
\label{app:protocol}

\subsection{Shared Evaluation Protocol}

The main evaluation uses Dana-i, insulin aspart, and real-time CGM for 72-hour episodes
under I2. The agent decides every 60 minutes with at most six event-triggered decisions
per local day. The shared prompt supplies the COGI-4-8 objective, personal therapy
settings, device limits, and available information. It prefers 5--6 mmol/L when safely
achievable; the scoring band remains 4--8 mmol/L. Each decision executes one environment
action. Personal notes persist within an episode and reset between tasks. All model backends request \texttt{high} reasoning effort. Table~\ref{tab:main212-harnesses}
lists the model harnesses.

\begin{table}[!htbp]
\centering\small
\caption{\textbf{Model harnesses, versions, and reasoning effort.}}
\label{tab:main212-harnesses}
\setlength{\tabcolsep}{6pt}
\begin{tabular*}{\linewidth}{@{\extracolsep{\fill}}lllc@{}}
\toprule
Model & Harness type & Version & Effort \\
\midrule
Claude Opus 5.5 & Claude Code & 2.1.296 & high \\
GPT 6.1 Sol & Codex CLI & 0.163.0 & high \\
DeepSeek 4.1 Flash & Claude Code & 2.1.296 & high \\
Kimi K3.1 Pro & Claude Code & 2.1.296 & high \\
GLM 5.3 Flash & Claude Code & 2.1.296 & high \\
\bottomrule
\end{tabular*}
\end{table}

\subsection{Baselines and Background Control}
\label{app:clinical}

\paragraph{BB and BB+Rules.}
We implement BB with a prescribed basal schedule and a bolus calculator
\citep{holt2021consensus,fox2020rl}. The calculator returns
$u=\max(0,fC/CR+q)$, where $C$ is reported carbohydrate intake, $CR$ is the carbohydrate
ratio, and $f=1$ for BB. Let $c=(G-G_{\mathrm{target}})/ISF$, where $G$ is observed glucose
and $ISF$ is insulin sensitivity. The correction term is $q=\max(0,c-IOB)$ for $c>0$
and $q=c$ otherwise; $IOB$ denotes nonnegative bolus insulin on board, with total insulin
on board as a fallback. Dose rounding, device limits, and minimum bolus intervals constrain
execution. BB also includes low-glucose treatment and responses to explicit device prompts.
Our BB+Rules extension adjusts $f$, incorporates CGM trends into $G$, and adds predefined
rules for exercise, illness, temporary basal changes, and device maintenance
\citep{riddell2017exercise}. Both use only visible reports and records, with 15-minute
consultation intervals in the main evaluation. Meal boluses require an available meal report.

\paragraph{OpenAPS.}
Our OpenAPS baseline uses the oref1 algorithm with npm release 0.7.1 of the official OpenAPS repository,
\url{https://github.com/openaps/oref0}, and its MIT-licensed dosing core~\citep{openaps},
with super-micro-boluses (SMB) and unannounced-meal (UAM) prediction enabled. Our adapter
constructs inputs from visible CGM, delivery records, personal therapy parameters, and
carbohydrate estimates, and enforces device limits. UAM deviation slopes and sensitivity
adaptation use approximate implementations.

\paragraph{OpenAPS decision frequency.}
We vary the fixed decision interval of this baseline from 15 to 240 minutes. Shortening
the interval from 240 to 15 minutes raises COGI from 56.84 to 71.19
(Table~\ref{tab:oref-cadence}), showing the benefit of more frequent feedback for this
controller.

\begin{table}[!htbp]
\centering\small
\caption{Decision-frequency ablation for OpenAPS. Entries are COGI, with higher values indicating better control.}
\label{tab:oref-cadence}
\begin{tabular*}{\linewidth}{@{\extracolsep{\fill}}lrrrrr@{}}
\toprule
Decision interval (min) & 15 & 30 & 60 & 120 & 240 \\
\midrule
COGI & 71.19 & 68.40 & 63.20 & 59.53 & 56.84 \\
\bottomrule
\end{tabular*}
\end{table}

\paragraph{HCL.}
HCL denotes the default DASH background controller: DASH-w/ CL enables it and DASH-w/o CL
disables it. This software layer follows public closed-loop designs and integrates the
OpenAPS core to adjust basal delivery when new CGM readings arrive~\citep{openaps}.
The agent retains its decision schedule and selects doses, carbohydrate intake, and
device actions; an agent-set temporary basal takes priority until expiry or cancellation.
The standalone reference runs the same closed-loop implementation every five minutes
under matching DASH/Fiasp, patient, seed, event, and CGM conditions, with rule-based
sensor maintenance and no agent.

\printTBD

\end{document}